\pdfoutput=1

\documentclass[11pt]{article}

\usepackage[]{naacl2021}

\usepackage{times}
\usepackage{latexsym}
\usepackage{booktabs}
\usepackage{graphicx}
\usepackage{todonotes}
\usepackage{adjustbox}
\usepackage{multirow}

\usepackage[T1]{fontenc}

\usepackage[utf8]{inputenc}

\usepackage{microtype}
\usepackage{xspace}

\newcommand{\FT}[0]{\includegraphics[width=.022\textwidth]{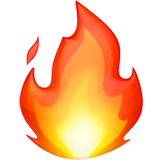}\xspace}
\newcommand{\SN}[0]{\includegraphics[width=.022\textwidth]{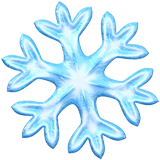}\xspace}

\title{%
Fine-Tuning Transformers for Identifying Self-Reporting Potential Cases and Symptoms of COVID-19 in Tweets}

\author{Max Fleming$^{1}$ Priyanka Dondeti$^{2}$  Caitlin N. Dreisbach$^{3}$  Adam Poliak$^{2,3}$ \\
 Johns Hopkins University$^{1}$ %
  Barnard College$^{2}$ \\ Data Science Institute, Columbia University$^{3}$\\
  \texttt{mflemi21@jhu.edu, \{pdd2112,apoliak\}@barnard.edu, c.dreisbach@columbia.edu} }

\begin{document}
\maketitle
\begin{abstract}
We describe our straight-forward approach for Tasks 5 and 6 of 2021 Social Media Mining for Health Applications (SMM4H) shared tasks. Our system is based on fine-tuning DistillBERT on each task, as well as first fine-tuning the model on the other task. %
We explore how much fine-tuning is necessary for accurately classifying tweets as containing self-reported COVID-19 symptoms (Task 5) or  whether a tweet related to COVID-19 is self-reporting, non-personal reporting, or a literature/news mention of the virus (Task 6).

\end{abstract}

\section{Introduction}
Fine-tuning off-the-shelf Transformer-based contextualized language models is a common baseline for contemporary Natural Language Processing~\cite{ruder2021lmfine-tuning}. When developing our system for %
\textbf{Task 6} of the 2021 Social Media Mining for Health Applications (SMM4H), we quickly discovered that fine-tuning DistilBERT~\cite{sanh2019distilbert}, a smaller and distilled version of BERT~\cite{devlin-etal-2019-bert}, outperformed training traditional, non-neural machine learning models.
Fine-tuning DistilBERT on the released training set %
resulted in a micro-F1 of $97.60$ on the Task 6 release development set. 
While this approach was not as successful for \textbf{Task 5} (binary-F1 of $51.49$),
in this paper, we explore how much fine-tuning is necessary for these tasks and whether there are benefits to first training the model on the other task since both are related to COVID-19.\footnote{All code developed is publicly available at
\url{https://github.com/mfleming99/SMM4H_2021}.}

\section{Task Description}
Both Task 5 and Task 6 focused on classifying tweets related to COVID-19~\cite{magge2021overview}. 
Task 5 required classifying tweets as describing self-reporting potential cases of COVID-19 or not. 
Tweets were extracted via manually crafted regular expressions for potential self-reported mentions of COVID-19 and then annotated by two people.
$1,148$ Tweets were labeled as containing a self-reporting potential cases  and $6,033$ were labeled as ``Other.'' The other tweets that might discuss COVID-19 but do not specifically reporting a user's or their household's potential cases were labeled as ``Other.''\footnote{See \newcite{info:doi/10.2196/25314} for a detailed description of the data collection and annotation protocols.}
Systems were ranked by F1-score for the “potential case” class.
\begin{table*}[t!]
    \centering
    \begin{tabular}{c|p{12cm}|c}
        \toprule
        Task & \multicolumn{1}{c|}{Tweet} & Label \\ \midrule
        & Just in case I do manage to contract \#coronavirus during the social distancing phase. I will kill it from the INSIDE! & Other \\
      \multirow{-2}{*}{Task5}  & So I've had this sore throat for a couple of days, I don't know if im being dramatic but i'm scared its Coronavirus?? & Potential \\ \midrule
      & New evidence suggests that neurological symptoms among hospitalized COVID-19 patients are extremely common &  Lit-News\\
     &   My dad tested positive for COVID-19 earlier this week, started having difficulty breathing this morning, and is now in the ED. & Nonpersonal \\
     \multirow{-5}{*}{Task6} &   Covid week 13 update. Week 11 kidney pain on the wane, presenting as high BP (affecting brain speed, vision, tightness in veins). & Self Report \\ 
        \bottomrule
    \end{tabular}
    \caption{Examples of tweets and labels for each task, abridged for space.} %
    \label{tab:data-example}
\end{table*}

In Task 6, systems must determine whether a tweet related to COVID-19 is self-reporting, non-personal reporting, or a literature/news mention of the virus. $1,421$ released examples are labeled as self-reporting, $3,567$ as non-personal reports, and $4,464$ as literature/news mentions.
Systems were evaluated by micro-F1 score. 
\autoref{tab:data-example} includes examples tweets from the development sets.

\section{Method} %
We fine-tuned DistillBERT using the implementation developed and released by HuggingFace transformer's library~\cite{wolf-etal-2020-transformers}. We trained the model for 3 epochs, using a batch size of 64 examples, warm-up steps of $500$ for the learning rate scheduler and a weight decay of $0.01$.
Following \newcite{peters-etal-2019-tune} recommendation to add minimal task hyper-parameters when fine-tuning pre-trained models, we used the remaining default hyper-parameters from the library's \texttt{Trainer} class.
All models were trained across 2 NVIDIA RTX 3090's.

\begin{figure*}[t!]
\centering
\includegraphics[width=0.49\textwidth,trim=0.5 0.5 10 10, clip]{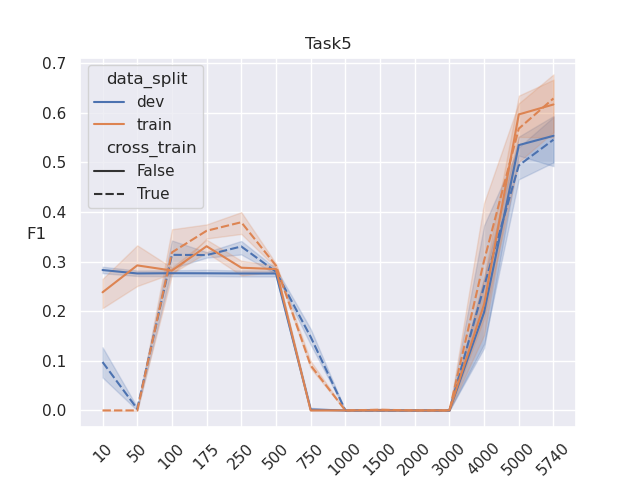}
\includegraphics[width=0.49\textwidth]{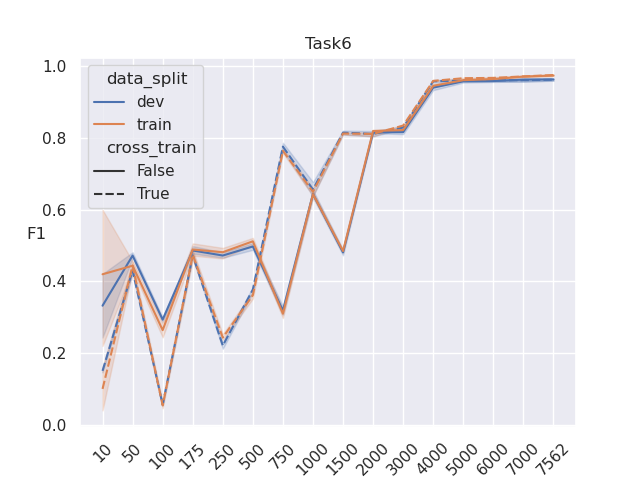}
\caption{$5$-fold results. The left and right graph respectively reflect binary-F1 results for Task 5 and micro-F1 results for Task 6. y-axes indicate F1 and x-axes indicate the number of training examples used. Dotted and solid lines, respectively, indicated that the model was pre-trained on the other task or not. Blue and orange respectively correspond to the training and development folds. The lines indicate the average across the $5$ folds and the shaded areas indicate the range of results.}
\label{fig:fine-tuning}
\end{figure*}

\subsection{Cross-validation}
We used $5$-fold evaluation to determine the utility of this simple approach.
For each task, we combined the training and development sets and removed duplicate tweets, resulting in $7,174$ and $9,452$ annotated examples for Task 5 and Task 6 respectively.\footnote{$7$ and $115$ examples were removed for Task 5 and 6 respectively.}  
We divided the datasets into 5 folds of roughly $1,435$ and $1,890$ labeled examples for Task 5 and Task 6 and fine-tune models on $4$ of the folds and test on the held out fold. For each fold, we fine-tuned the model on a increasing number of training examples: $10$, $50$, $100$, $175$, $250$, $500$, $750$, $1$K, $1.5$K, $2$K, $3$K, $4$K, $5$K, $6$K, $7$K, $8$K.\footnote{For Task 5, the maximum number of training examples are $5,740$} 
Additionally, for both tasks, we experimented with using a model pre-trained on the other task. We hypothesized this might be beneficial as these tasks seem to be related.

\section{Results}
\autoref{fig:fine-tuning} shows the results of fine-tuning DistillBert on each task. For Task 5 (left graph), when fine-tuning on $50$ examples or less, initially training on Task 6 (dotted lines) is detrimental.
When fine-tuning on somewhere between $50$ and $100$ training examples, first training the models on Task 6 leads to a noticeable improvement. This continued until we fine-tuned the model on $500$ examples. Once we fine-tuned the model on $1000$ to $3000$ examples, there is no difference between first training on the other task as the models only predict the majority class ``Other". As the number of training examples increases from this point, we begin to see large improvements and larger variances between the models trained on different folds. First training on Task 6 appears to be most beneficial when fine-tuning on $100$ through $750$ Task 5 examples.

For Task 6 (right graph), the benefits of pre-training the model on Task 5 are not as clear cut, and the results oscillate a bit more. It seems that pre-training on Task 5 is only beneficial when fine-tuning the model on $750$ through $2,000$ examples (except for the case when fine-tuning on $1,000$ examples). For both tasks, pre-training on the other task seems to make no difference once the model is fine-tuned on enough task specific examples (roughly $1,000$ and $2,000$ examples for Task 5 and Task 6).

\paragraph{Held out test set}
In these experiments, the model performance on the held out fold seems to increase as we add more training examples. While results for Task 6 seem to plateau, we notice a small increase as we continue to add training examples. Therefore, for our official submissions, we fine-tuned the model on all released examples.

\autoref{tab:test-results} reports results for the official test sets.\footnote{These numbers differ from the official leaderboard during the evaluation as we discovered a bug related to loading our pre-trained models during the post-evaluation period.}
The $63.19$ binary-F1 for Task 5 might indicate that training on more examples is beneficial for this task. For Task 6, we notice the micro-F1 drops a bit compared to the results on the held out folds. 
For both tasks, pre-training on the other task is not beneficial on the test set when trained on as many labeled examples as possible.

We also include a majority vote ensemble of the $5$-fold models trained on different training sizes. These test results follow the general trends in \autoref{fig:fine-tuning} indicating when it is most beneficial to first train the DistilBert model on the other task. 
Similar to the results in \autoref{fig:fine-tuning}, when fine-tuning on $750$ through $3,000$ Task 5 examples, the model achieved a $0$ binary-F1 since it always predicted the majority class ``Other.''

\begin{table}[t!]
    \centering
    \begin{tabular}{c|cc|cc}
    \toprule
        Train Size & \multicolumn{2}{c}{Task5} & \multicolumn{2}{c}{Task6} \\
        & \SN & \FT & \SN & \FT \\
        \midrule
        -- &	63.19  & 62.24 & 92.88 & 91.77\\ \midrule
        50 & 29.33 & 05.72 & 46.17 & 42.75 \\
        100 & -- & -- & 27.65 & 05.72\\
        175 & 28.54 & 32.22 & 47.97 & 46.20\\
        250 & -- & -- & 46.15 & 36.83\\
        500 & 29.29 & 28.92 & - & -\\
        750 & 00.00 &  16.00 & 31.02 & 56.70\\
        1000 & 00.00 & - & - & - \\
        2000 & - & - & 80.11 & - \\
        4000 & - & - & 92.41 & - \\
        5000 & 55.69 & 51.19 & - & - \\

         \bottomrule
        
    \end{tabular}
    \caption{Results on the official test sets available on CodaLabs. Numbers indicate binary-F1 for Task 5 and micro-F1 for Task 6.
    \SN indicates the model was fine-tuned on the specific task and \FT indicates the model was first fine-tuned on the other task. The first line reports the results trained on the combination of the corresponding train and development sets - $7,174$ for Task 5 and $9,452$ for Task 6. The remaining lines are based on a ensemble of the $5$ models trained on the corresponding number of examples using a majority vote.}
    \label{tab:test-results}
\end{table}

\section{Conclusion}
We discussed our straightforward approach of fine-tuning a DistilBert model on Tasks 5 and 6 of the 2021 Social Media Mining for Health Applications shared tasks. While not attaining state-of-the-art, these results are competitive and demonstrate the benefit of leveraging large scale pre-trained contextualized language models. We additionally explored the benefits of first training the model on the corresponding task and determine when this can be beneficial. Future work might consider jointly fine-tuning a Bert-based model on both tasks using a multi-task approach as opposed to the transfer learning approach employed here.

\section*{Acknowledgements}
We would like to thank the
anonymous reviewer for their feedback. Our experiments were conducted using computational infrastructure provided by the Barnard Vagelos Computational Science Center.

\bibliography{references}

\begin{thebibliography}{7}
\expandafter\ifx\csname natexlab\endcsname\relax\def\natexlab#1{#1}\fi

\bibitem[{Devlin et~al.(2019)Devlin, Chang, Lee, and
  Toutanova}]{devlin-etal-2019-bert}
Jacob Devlin, Ming-Wei Chang, Kenton Lee, and Kristina Toutanova. 2019.
\newblock \href {https://doi.org/10.18653/v1/N19-1423} {{BERT}: Pre-training of
  deep bidirectional transformers for language understanding}.
\newblock In \emph{Proceedings of the 2019 Conference of the North {A}merican
  Chapter of the Association for Computational Linguistics: Human Language
  Technologies, Volume 1 (Long and Short Papers)}, pages 4171--4186,
  Minneapolis, Minnesota. Association for Computational Linguistics.

\bibitem[{Klein et~al.(2021)Klein, Magge, O'Connor, Flores~Amaro,
  Weissenbacher, and Gonzalez~Hernandez}]{info:doi/10.2196/25314}
Ari~Z Klein, Arjun Magge, Karen O'Connor, Jesus~Ivan Flores~Amaro, Davy
  Weissenbacher, and Graciela Gonzalez~Hernandez. 2021.
\newblock \href {https://doi.org/10.2196/25314} {Toward using twitter for
  tracking covid-19: A natural language processing pipeline and exploratory
  data set}.
\newblock \emph{J Med Internet Res}, 23(1):e25314.

\bibitem[{Magge et~al.(2021)Magge, Klein, Flores, Alimova, Al-garadi,
  Miranda-Escalada, Miftahutdinov, Farr{\'e}-Maduell, L{\'o}pez, Banda,
  O’Connor, Sarker, Tutubalina, Krallinger, Weissenbacher, and
  Gonzalez-Hernandez}]{magge2021overview}
Arjun Magge, Ari Klein, Ivan Flores, Ilseyar Alimova, Mohammed~Ali Al-garadi,
  Antonio Miranda-Escalada, Zulfat Miftahutdinov, Eul{\`a}lia
  Farr{\'e}-Maduell, Salvador~Lima L{\'o}pez, Juan~M Banda, Karen O’Connor,
  Abeed Sarker, Elena Tutubalina, Martin Krallinger, Davy Weissenbacher, and
  Graciela Gonzalez-Hernandez. 2021.
\newblock Overview of the sixth social media mining for health applications (\#
  smm4h) shared tasks at naacl 2021.
\newblock In \emph{Proceedings of the Sixth Social Media Mining for Health
  Applications Workshop \& Shared Task}.

\bibitem[{Peters et~al.(2019)Peters, Ruder, and Smith}]{peters-etal-2019-tune}
Matthew~E. Peters, Sebastian Ruder, and Noah~A. Smith. 2019.
\newblock \href {https://doi.org/10.18653/v1/W19-4302} {To tune or not to tune?
  adapting pretrained representations to diverse tasks}.
\newblock In \emph{Proceedings of the 4th Workshop on Representation Learning
  for NLP (RepL4NLP-2019)}, pages 7--14, Florence, Italy. Association for
  Computational Linguistics.

\bibitem[{Ruder(2021)}]{ruder2021lmfine-tuning}
Sebastian Ruder. 2021.
\newblock {Recent Advances in Language Model Fine-tuning}.
\newblock \url{http://ruder.io/recent-advances-lm-fine-tuning}.

\bibitem[{Sanh et~al.(2019)Sanh, Debut, Chaumond, and
  Wolf}]{sanh2019distilbert}
Victor Sanh, Lysandre Debut, Julien Chaumond, and Thomas Wolf. 2019.
\newblock Distilbert, a distilled version of bert: smaller, faster, cheaper and
  lighter.
\newblock \emph{arXiv preprint arXiv:1910.01108}.

\bibitem[{Wolf et~al.(2020)Wolf, Debut, Sanh, Chaumond, Delangue, Moi, Cistac,
  Rault, Louf, Funtowicz, Davison, Shleifer, von Platen, Ma, Jernite, Plu, Xu,
  Le~Scao, Gugger, Drame, Lhoest, and Rush}]{wolf-etal-2020-transformers}
Thomas Wolf, Lysandre Debut, Victor Sanh, Julien Chaumond, Clement Delangue,
  Anthony Moi, Pierric Cistac, Tim Rault, Remi Louf, Morgan Funtowicz, Joe
  Davison, Sam Shleifer, Patrick von Platen, Clara Ma, Yacine Jernite, Julien
  Plu, Canwen Xu, Teven Le~Scao, Sylvain Gugger, Mariama Drame, Quentin Lhoest,
  and Alexander Rush. 2020.
\newblock \href {https://doi.org/10.18653/v1/2020.emnlp-demos.6} {Transformers:
  State-of-the-art natural language processing}.
\newblock In \emph{Proceedings of the 2020 Conference on Empirical Methods in
  Natural Language Processing: System Demonstrations}, pages 38--45, Online.
  Association for Computational Linguistics.

\end{thebibliography}
\bibliographystyle{acl_natbib}

\appendix

\end{document}